\documentclass[conference]{IEEEtran}
\usepackage{graphicx}
\usepackage{algorithm}
\usepackage{algpseudocode}
\usepackage[caption=false,font=footnotesize]{subfig}

\begin{document}
\title{Enhancing Object Detection Accuracy in Autonomous Vehicles Using Synthetic Data}



\author{\IEEEauthorblockN{Sergei Voronin}
\IEEEauthorblockA{Wellington Institute of Technology\\
Wellington, New Zealand
}
\and
\IEEEauthorblockN{Abubakar Siddique}
\IEEEauthorblockA{Wellington Institute of Technology\\
Wellington, New Zealand
}
\and
\IEEEauthorblockN{Muhammad Iqbal}
\IEEEauthorblockA{Higher Colleges of Technology\\
Fujairah, United Arab Emirates
}
}

\maketitle

\begin{abstract}
The rapid progress in machine learning models has significantly boosted the potential for real-world applications such as autonomous vehicles, disease diagnoses, and recognition of emergencies. The performance of many machine learning models depends on the nature and size of the training data sets. These models often face challenges due to the scarcity, noise, and imbalance in real-world data, limiting their performance. Nonetheless, high-quality, diverse, relevant and representative training data is essential to build accurate and reliable machine learning models that adapt well to real-world scenarios.

It is hypothesised that well-designed synthetic data can improve the performance of a machine learning algorithm. This work aims to create a synthetic dataset and evaluate its effectiveness to improve the prediction accuracy of object detection systems. This work considers autonomous vehicle scenarios as an illustrative example to show the efficacy of synthetic data. The effectiveness of these synthetic datasets in improving the performance of state-of-the-art object detection models is explored. The findings demonstrate that incorporating synthetic data improves model performance across all performance matrices.

Two deep learning systems, \textit{System-1} (trained on real-world data) and \textit{System-2} (trained on a combination of real and synthetic data), are evaluated using the state-of-the-art YOLO model across multiple metrics, including accuracy, precision, recall, and mean average precision. Experimental results revealed that \textit{System-2} outperformed \textit{System-1}, showing a $3\%$ improvement in accuracy, along with superior performance in all other metrics. 
\end{abstract}

\begin{IEEEkeywords}
Object Detection, Deep Learning, Synthetic Data, Autonomous Vehicle.
\end{IEEEkeywords}

\IEEEpeerreviewmaketitle

\section{Introduction}
\label{Intro}
The rapid technological advancements in machine learning enable automatic data processing, diagnosis, and recognition of emergencies. The number of applications and services enhancing vehicle handling, road safety, and identifying and monitoring potentially hazardous objects has grown yearly. Applications of machine learning (ML) range from simple statistical displays to complex integrations with autopilot systems, where traffic is monitored based on incoming data at runtime. Failures in these integrated systems can cause significant damage to both vehicles and passengers. 

High-quality, diverse, relevant and representative training data is essential to build accurate and reliable machine learning models that adapt well to real-world scenarios. Noise is an integral part of real-world datasets that can mislead the learning process of machine learning models \cite{siddique2023modern,siddique2020lateralized}. In emergency transport situations, real-world data may not be readily available for immediate feedback and updated visualisations due to the dynamic nature of the events and the priority of providing timely medical care over data collection. 

Many ML models require a human-in-the-loop for comprehensive training for learning complex real-world tasks such as the same disease may look different from various angles and conducting high-quality shooting in special conditions like deep water can be quite challenging. Moreover, some models show high sensitivity to lighting and noise conditions, which reduces the accuracy and performance of the obtained materials \cite{saleh2022computer,siddique2021lateralizedthesis}. 

Another common challenge for ML models while solving real-world tasks is the imbalance in the number of examples for different categories within the training dataset \cite{lu2020identifying,siddique2023lateralized}. Many real-world datasets, especially those related to diseases, are naturally imbalanced. This imbalance in the training samples can affect the learning of ML models, making it more focused on the groups with many examples. Consequently, the trained models may have issues like wrongly identifying rare classes, such as fish species, patients with brain stroke, etc. The simplest way to fix these problems is to create training data with more balanced groups \cite{ju2024survey,spelmen2018review}.

Another important aspect of data is its diverse sources. Training datasets cannot capture all possible variations present in the real world. For instance, in the task of determining vehicle damage for insurance claims, data can come from various sources, such as smartphone cameras, outdoor surveillance cameras, and professional photographers' cameras. This results in differences in photo quality, lighting, and perspective. Images may often contain too many objects or have excessively high resolution. Additionally, training object-detection models require a large volume of labelled data. For instance, in the case of plant disease classes, it can be challenging to gather a large amount of video and information, as some diseases are rare and practically lead to the immediate death of the plant \cite{mrisho2020accuracy}.

\begin{figure}
	\begin{center}
		\includegraphics [scale=0.5]{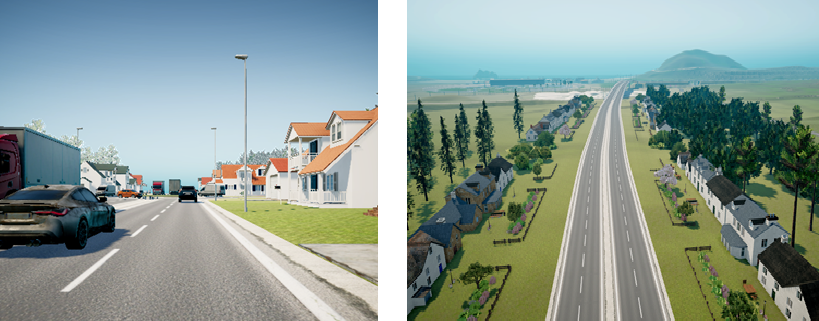}
		\caption{Sample synthetic images.}
		\label{SampleImgs}
	\end{center}
\end{figure}

A promising solution to address the above-mentioned problems is synthetic data. Synthetic data is artificially generated data that mimics the statistical properties and patterns of real-world data. It is created through computational methods, simulations, and ML techniques \cite{nikolenko2021synthetic}. Two sample synthetic images are shown in Fig. \ref{SampleImgs}. Synthetic data have been successfully used to improve the performance of ML models. For instance, ML models for the biomedical domain utilize synthetic data to supplement the scarce real-world dataset, resulting in a comprehensive collection of medical images depicting various patient conditions \cite{dahmen2019synsys}.

The use of synthetic data offers additional benefits, such as controlling the appearance, lighting, and background of objects or scenes through handcrafted or engine-based rendering processes \cite{johnson2016driving}. This level of control is precious when simulating complex or rare scenarios that are difficult to capture in real-world datasets. Moreover, synthetic data can provide accurate labels without extensive manual annotation, enabling large-scale data generation \cite{chen2021synthetic}.

The lack of suitable real-world data can hinder the ability to make informed decisions and allocate resources effectively during critical incidents. Synthetic data facilitates the creation of detailed storyboards for rapidly dangerous situations and the application of specifications for decision-making. Synthetic data is also useful when real data is inaccessible, lost, highly confidential, insufficiently detailed, of the wrong scale, or incomplete. Furthermore, creating a system that performs reliably during the testing and future predictions imposes specific requirements on data such as diverse scenarios and convergence under varying weather conditions. In such cases, the use of synthetic data, where these parameters are carefully controlled and configurable, can be an effective solution unaffected by external factors \cite{paulin2023review}

It is hypothesised that well-designed synthetic data can improve the performance of a machine learning algorithm. The main goal of this work is to create a synthetic dataset and evaluate its effectiveness to improve the prediction accuracy of object detection systems. This work considers autonomous vehicle scenarios as an illustrative example to show the efficacy of synthetic data. To create an environment similar to real-world situations, 3D scenes of roadbeds and low-rise settlements are needed. Fine-grained image textures from these scenarios are required to facilitate the training of an object detection system and provide accurate depth information.



\section{Background}
\label{Bckgrnd}
The goals of this section are two-fold: first, to introduce the relevant terminologies and techniques that will inform the work; second, to review the current state-of-the-art techniques that have been developed to address object detection problems.

\subsection{Deep neural networks}
Deep neural networks have been commonly used to solve a wide range of real-world problems ranging from autonomous vehicles to disease detection \cite{saleh2022computer}. Deep networks extract valuable features from the raw images, independently discover hidden patterns, and progressively learn various levels of complexity \cite{alzubaidi2021review}. Deep networks need large training data sets, depending on the nature and complexity of the problem, to learn patterns and hidden features. Real-world data is often expensive, noisy, limited, and imbalanced. This work aims to generate synthetic data and evaluate its effectiveness in improving the performance of ML models.

Deep network-based autonomous driving systems have been used to improve performance and enhance safety by detecting obstacles, recognising road signs, tracking routes, and identifying pedestrians and cyclists \cite{chen2017multi}. The single shot detector (SSD) is a well-known real-time object detection method based on deep convolutional neural networks (CNN). CNN processes an input image and generates a set of bounding boxes and class predictions for each detected object. It combines localisation and classification in a single forward pass through the network. It uses a series of convolutional layers with decreasing spatial resolution to detect objects at different scales and aspect ratios. SSD is designed to be fast and efficient, making it ideal for real-time applications like video surveillance and autonomous driving. Representatives of single-stage detectors include models like YOLO, SSD, and RetinaNet. 


One of the best machine learning models for object detection is the YOLO (\textit{You Only Look Once}) model, which has gained widespread popularity due to its exceptional speed and high accuracy. The YOLO model employs a single neural network to simultaneously predict bounding boxes and class probabilities for objects in an image. In YOLO-based systems, the essence of the target detection algorithm lies in its compact model size and rapid computation speed \cite{jiang2022review}. YOLO's design is simple, allowing it to output the positions and categories of bounding boxes directly. Its speed advantage comes from the fact that YOLO reformulates the target detection problem as a regression problem and it only requires inputting the image into the network to obtain the final detection result, enabling real-time video detection. By using the entire image for detection, YOLO can encode global information and reduce errors where the background is mistakenly identified as an object. YOLO model will be used for conducting experiments in this work. 

\subsection{Realtime graphic engines}
Real-time graphic engines, such as Unity and Unreal Engine, revolutionise visual experiences by rendering lifelike graphics instantaneously, transforming how we interact with digital environments \cite{borkman2021unity}. Unity has been commonly used to create various virtual environments to enhance machine learning models' learning performance. It offers versatility in achieving realistic visuals and simulations and user-friendly tools for programmers. Unity is designed for both 3D and 2D games, serving as an excellent tool for simulating most commercial products and platforms. This work will use the Unity engine to create environments and generate synthetic data.

\section{Procedural Data Generation}
A significant challenge in machine learning is procedural data generation (PDG), which deals with the difficulty or lack of real-world data acquisition for model training. PDG offers an automated approach to generating synthetic data with high accuracy and precision. This is especially useful when gathering empirical data is costly, time-consuming, or morally dubious. It is critical to design complex urban and natural environments to effectively educate AI systems in handling a wide range of real-world scenarios \cite{shah2018airsim}. Real-world events must be simulated to evaluate whether a system can handle and react to unexpected circumstances. Before AI models are applied in crucial real-world scenarios, such as spaceship control systems or disaster response robots, PDG offers a controlled and flexible data environment that enables extensive testing and optimisation.

The Unity framework has been commonly used to generate terrain and procedurally place objects such as trees and rocks. Procedural generation frameworks, which streamline the production of synthetic data, have become valuable tools for improving the simulation of different scenarios. This enhances the thoroughness of testing, increasing the reliability of AI systems. These frameworks facilitate the simulation of rare or exceptional situations, ultimately enhancing AI’s ability to respond to unforeseen events. This capability significantly improves the safety and efficiency of ML-based systems when applied in dynamic real-world environments where unexpected occurrences are inevitable \cite{richter2016playing}.

In conclusion, procedural data generation effectively addresses the challenge of limited real-world data by producing high-quality synthetic data, enabling the development of more robust and flexible machine learning models. As this approach gains importance across various industries, it is accelerating the advancement of more accurate AI-driven models for autonomous vehicles.

\section{Environment Configuration}
The Unity environment is configured to efficiently manage 3D scenes using parent-child object dependencies, where child objects inherit transformations from their parent. Textures, which wrap around 3D models, should have dimensions in powers of two for optimal rendering performance. Materials and shaders are applied to enhance lighting and visual effects. Unity’s Level of Detail (LOD) system reduces model complexity as objects move farther from the camera, improving performance without sacrificing visual quality, while occlusion culling prevents rendering objects outside the camera’s view. 

Unity’s plugin architecture allows developers to extend the functionality of the engine with tools tailored to specific needs. In this work, several key plugins are used to enhance the environment configuration and overall simulation:

\subsection{EasyRoads3D}
This plugin is used to create realistic road networks, including highways and traffic elements. It streamlines the process of designing curved roads, adding road markings, and placing objects like trees and traffic signs, enabling rapid development of detailed road environments.

\subsection{Mobile Traffic System}
To simulate real-world road traffic, this plugin is employed to manage the movement of vehicles within the scene. It supports the modelling of various traffic scenarios such as intersections, highways, and interchanges while considering factors like traffic density, road signs, and driver behaviour. The system allows for dynamic adjustment of traffic conditions, including speed limits, vehicle types, and environmental factors such as weather and time of day.

\subsection{GAIA}
GAIA is a powerful terrain generation tool used to create detailed and optimized landscapes. The terrain creation process begins with terrain stamping, where features like mountains, valleys, and rivers are procedurally generated. GAIA then applies realistic textures based on terrain properties like slope and height, and populates the scene with vegetation and other objects according to biome-specific rules. The plugin’s ability to optimize object meshes and textures ensures that the scene runs smoothly even on low-end devices.

\subsection{Pack of Modular Homes}
The Pack of Modular Homes (PMH) plugin is used to generate modular houses for scene development, particularly in village creation. The plugin includes a comprehensive collection of modular building elements, textures, and materials, enabling the rapid construction of diverse building types with varying designs. This modular approach allows for efficient scene expansion by assembling custom houses from a set of predefined components, resulting in highly diverse and realistic settlements.

By leveraging Unity’s game object hierarchy, texture management, and plugin ecosystem, the environment is configured to support a realistic and efficient simulation. The combination of procedural generation tools, LOD management, and specialized plugins allowed for the creation of complex scenes that are both visually rich and performance-optimized.

\begin{figure}
	\begin{center}
		\includegraphics [scale=0.5]{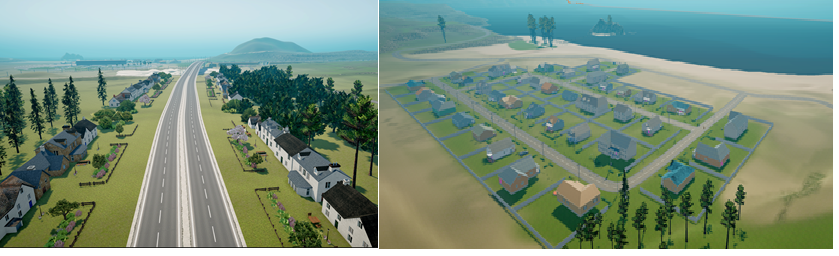}
		\caption{Sample Images from the Settlement Complex scene}
		\label{ImgsScene2}
	\end{center}
\end{figure}

\section{Environment Creation}
A series of objects with specific characteristics are used to generate a set of synthetic data in the scene, which allows recreating a 3D scene similar to the terrain of a real-world dataset. The details of two sample scenes that are created to model locations similar to real-world environments are presented in the following sections.

\subsection{Settlement Complex} 
\begin{figure}
	\begin{center}
		\includegraphics [scale=1.1]{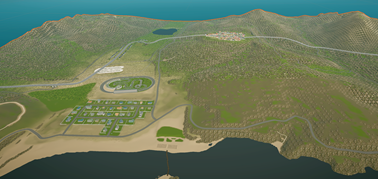}
		\caption{Settlement Complex Scene}
		\label{SettlementComplex}
	\end{center}
\end{figure}

The purpose of this scene is to create a diverse anthropogenic landscape that integrates residential settlements, industrial areas, and various road networks. The scene features agricultural lands with a linear arrangement of residential settlements along a major road. One settlement is located near a pond, another by the sea, and a third on a hill. Additionally, an industrial complex is positioned near the coastline. Several medium-sized roads traverse the landscape, extending through mountainous areas, valleys, and along the ocean, offering a comprehensive view of the terrain (see Fig. \ref{ImgsScene2}. By incorporating these anthropogenic elements into synthetic datasets, the training data becomes more representative of the varied landscapes and scenarios that self-driving cars will encounter in the real world. A snapshot of the Settlement Complex scene is shown in Fig. \ref{SettlementComplex}.

\subsection{Mountain Lowland}
\begin{figure}
	\begin{center}
		\includegraphics [scale=1.1]{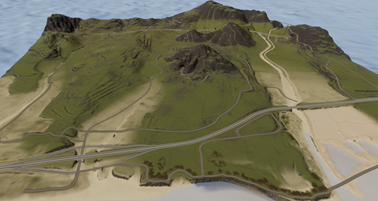}
		\caption{Mountain Lowland Scene}
		\label{MountainLowland}
	\end{center}
\end{figure}

The main purpose of this scene is to create a natural terrain with minimal or no man-made structures. For instance, there are no buildings, only different types of roads, including high-speed ones with concrete barriers and highways. The Mountain Lowland scene is designed to provide an environment that closely resembles the real-world traffic scenarios in New Zealand. A sample snapshot of the Mountain Lowland scene is shown in Fig. \ref{MountainLowland}.

\begin{figure}
	\begin{center}
		\includegraphics [scale=1.2]{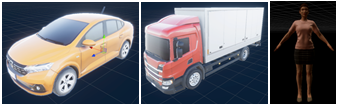}
		\caption{Target class models}
		\label{TargetClasses}
	\end{center}
\end{figure}

In both the \textit{Settlement Complex} and \textit{Mountain Lowland} scenes, a variety of objects are used to create realistic and dynamic environments, including vehicles, people, buildings, greenery, roads, and natural landscapes featuring water bodies. The target classes for machine learning models include cars, people, and trucks (see Fig. \ref{TargetClasses}), while non-target classes comprise the landscape, greenery, water bodies, roads, buildings, and other landscaping elements like lamp posts, mailboxes, and haystacks. The interaction rules between these objects are programmed into the simulation, ensuring a cohesive scenario. Additionally, several buildings were augmented to introduce colour variation.

\subsection{Object Behavior}
There are two types of objects in the scenes: static objects and dynamic objects. Static objects are non-moving elements that remain fixed in their positions within the scene. They do not have any animation or interaction with other objects. These include landscapes, buildings, roads, greenery, water bodies, and road infrastructure. Static objects serve as the foundational backdrop of the environment, providing the necessary context and realism to create an immersive setting.

In contrast, dynamic objects can move, animate, or interact with other objects within the scene. Dynamic objects include vehicles such as cars and trucks, as well as pedestrians. Vehicles are animated to move in straight lines or follow curved trajectories, with the ability to turn left or right at intersections. Speed variations, including acceleration and deceleration, are also supported. While collisions between vehicles or with static objects can occur, the deformation of objects upon impact is not simulated. Pedestrians are animated for slow, uniform movement, either walking along straight paths or crossing at designated pedestrian crossings.

\section{Data Set Generation}
\begin{figure}
	\begin{center}
		\includegraphics [scale=1]{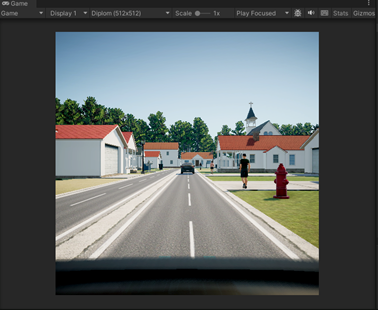}
		\caption{A sample view from a car}
		\label{CarView}
	\end{center}
\end{figure}

To generate the synthetic dataset, the Unity engine is used to replicate real-world road situations, ensuring a balanced representation of different object classes, such as vehicles, pedestrians, and trucks. A total of 10,000 images are created using Unity’s \textit{Perception Camera} component. This tool automatically labels target class objects (cars, people, trucks) by attaching a bounding box to each object and saving the corresponding metadata. The metadata, initially stored in SOLO format, was converted into the YOLO format using a custom script to meet the requirements of the YOLO model, which will be used for object detection. The dataset is recorded from a camera placed on the front of a randomly selected car (see Fig. \ref{CarView}, ensuring diversity in the generated data for each run. The resolution of each image is 512x512, matching that of the base dataset to enable seamless integration into the final training datasets. 

This work aims to demonstrate that using synthetic data can improve the performance of a machine-learning model in addressing real-world problems. There are various methods to utilise synthetic data for this purpose. One straightforward approach is to train a model solely on synthetic data and then apply it to real images. However, it is widely recognised that classification models trained exclusively on synthetic images frequently need domain adaptation to perform effectively on real-world images \cite{sun2014virtual}. This necessity arises because virtual and real-world cameras are different in nature. Consequently, domain adaptation is frequently needed even when the training and testing images originate from different real-world camera sensors \cite{vazquez2013virtual}. 

The well-known Balanced Gradient Contribution (BGC) method is used for this work \cite{ros2016training}. This technique involves creating batches with images for both virtual and real-world domains in a specific ratio. The majority of the data distribution consists of real-world images, while the synthetic images play an essential role in enhancing the model's ability to generalize and prevent overfitting. As a result, the model is trained by considering the statistical properties and characteristics from both the real-world and synthetic data domains. This comprehensive training approach leads to the development of a model that demonstrates strong performance across both types of data.

A well-known and publicly available real-world data set \textit{Self-Driving Cars} is selected from Kaggle. This dataset contains a large collection of front-view images captured from a car while driving. The images cover a wide range of road scenarios, including urban and highway environments, with varying lighting conditions and potential obstacles, providing a realistic and diverse training set for autonomous vehicle systems.

To evaluate the effectiveness of the synthetic data, three types of datasets are created, i.e., test dataset, training dataset 1, and training dataset 2. The test dataset consists of $914$ randomly selected real-world images. Training dataset 1 contains $4658$ real-world images and is used exclusively to train System-1. Training dataset 2 is a combination of real-world and synthetic images. It includes the same $4658$ real-world images from training dataset 1, along with $2139$ synthetic images generated using Unity. This dataset is used to train System-2. This setup enables a direct comparison of the performance of models trained on purely real-world data versus those trained on a combination of real and synthetic data.

\section{Experimental Evaluation}
Two deep learning systems, named System-1 and System-2, are used for the evaluation of the proposed approach. Both the systems are trained for 200 epochs using the same machine learning model YOLO. The System-1  runs with training dataset 1, which contains only real data, and the System-2 runs with training dataset 2, which contains both real and synthetic data. Both systems are tested on the same test dataset. The training data set is further divided into training and validation data at the ratio 80:20, respectively. All real and synthetic images are resized to the same resolution of 512 × 512. Moreover, all the experiments are repeated ten times using different seed values and the results presented here are the average from those runs.  

During training, the batch size determines how many images are used in one iteration to calculate the loss and update the model's weights. For System-1, each batch contains only real-world images. However, for System-2, each batch contains 5 images from the real domain and 3 from the synthetic domain using the Balanced Gradient Contribution method to ensure an even contribution from both domains during model optimization. The performance of the deep learning models, System-1 and System-2, is assessed using several metrics, including accuracy, precision, recall, mean average precision, and F1 score. 

The experimental results of both systems are shown in Table \ref{ExpResTable}. System-1 achieved an accuracy of $0.57$, with a precision of $77.46\%$ and a recall of $58.06\%$. Its mean average precision is $64.50\%$, resulting in an F1 score of $0.662$. In contrast, System-2 demonstrated improved performance across all metrics, achieving an accuracy of $0.60$. It recorded a precision of $82.56\%$ and a recall of $61.71\%$, with a mean average precision of $70.37\%$ and an F1 score of $0.705$. These results indicate that System-2, which is trained on a combination of real and synthetic data, outperformed System-1, which utilized only real-world data. The enhancements in precision, recall, and overall F1 score suggest that incorporating synthetic data positively impacts the model's ability to recognize and classify objects effectively.

\begin{table}[!t]
	\caption{Performance Evaluation}
        \label{ExpResTable}
	\resizebox{\columnwidth}{!}{%
    	  \begin{tabular}{|c|c|c|c|c|c|}
                \hline
                         & \textbf{Accuracy} & \textbf{Precision} & \textbf{Recall} & \textbf{F1 Score} & \textbf{Mean Average Precision} \\ \hline
                System-1 & 0.57              & 77.46\%            & 58.06\%         & 0.662             & 64.50\%                         \\ \hline
                System-2 & 0.60              & 82.56\%            & 61.71\%         & 0.705             & 70.37\%                         \\ \hline
                \end{tabular}
        }
	\vspace{-1.20em}
\end{table}

To analyze the learning pattern of both models, three types of loss functions are employed in this study. First, box loss, quantifies how well the model predicts the location and size of the bounding boxes around the objects. Lower box loss means the model is better at predicting where objects are and their scale. The second, classification loss ($cls\_loss$), measures the accuracy of predicting the correct category of the objects within the bounding boxes. Lower $cls\_loss$ points towards a more accurate classification of objects. The third loss function, distribution focal loss ($dfl\_loss$), is designed to address the class imbalance by down-weighting easy examples and focusing the training on hard-to-classify objects. It's a modified version of focal loss used in object detection tasks, particularly when some classes are rarer than others. 

\begin{figure}
	\begin{center}
		\includegraphics [scale=0.68]{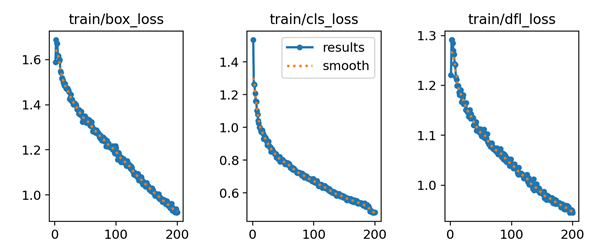}
		\caption{A sample graph of loss functions}
		\label{LossFuncs}
	\end{center}
\end{figure}

A sample graph of loss functions is shown in Fig. \ref{LossFuncs}. Box loss decreased from $1.7$ to $0.95$ in $200$ epochs, indicating that the model has improved in predicting object locations and sizes. The classification loss exhibited a sharp decline from $1.5$ in the first $20$ epochs, eventually stabilizing around $0.5$, which demonstrates the model's enhanced capability to recognize object categories within bounding boxes. Meanwhile, the distribution focal loss initially rose to nearly $1.3$ before dropping to $0.95$, reflecting the model's progress in distinguishing similar objects effectively.

\section{Conclusion}
This work successfully demonstrated the effectiveness of using synthetic data to enhance the performance of state-of-the-art machine learning models for object detection tasks. The synthetic data generation process effectively addressed the issue of object diversity by varying background elements and textures, resulting in a dataset that closely mirrors real-world distributions. The experimental results clearly showed that System-2, which incorporated synthetic data, consistently outperformed System-1 across all performance matrices.

This study highlights the potential of synthetic data generation in overcoming the challenges of limited real-world datasets, especially in fields like healthcare, where large, balanced, and privacy-safe datasets are often difficult to obtain. By creating diverse, synthetic datasets that include rare cases, researchers can build more robust and generalizable models while addressing privacy concerns.

\bibliographystyle{IEEEtran}
\bibliography{BibAB4Trans}


\end{document}